\documentclass[lettersize,journal]{IEEEtran}
\usepackage{amsmath,amsfonts}
\usepackage{algorithm}
\usepackage{algorithmicx}
\usepackage{algpseudocode}
\usepackage{array}
\usepackage{textcomp}
\usepackage{stfloats}
\usepackage{url}
\usepackage{verbatim}
\usepackage{graphicx}
\usepackage{cite}
\usepackage{csquotes}
\usepackage{rotating}
\usepackage{multirow}
\usepackage{times}
\usepackage{epsfig}
\usepackage{color}
\usepackage[normalem]{ulem}
\usepackage{booktabs}
\usepackage{subfigure}

\usepackage{bbm}

\usepackage[colorlinks=true,linkcolor=black,citecolor=blue,urlcolor=blue,]{hyperref}

\newcommand\eg{\emph{e.g.}}

\newcommand\wrt{w.r.t.}

\begin{document}

\title{Semi-supervised Salient Object Detection with Effective Confidence Estimation}


\author{Jiawei Liu,~
        Jing Zhang,~
        Nick Barnes\\
\IEEEcompsocitemizethanks{
\IEEEcompsocthanksitem Jiawei Liu, Jing Zhang, Kaihao Zhang and Nick Barnes are with School of Computing, the Australian National University, Canberra, Australia ({jiawei.liu3, jing.zhang, kaihao.zhang, nick.barnes}@anu.edu.au).
}

}

\markboth{Journal of \LaTeX\ Class Files,~Vol.~14, No.~8, August~2021}%
{Shell \MakeLowercase{\textit{et al.}}: A Sample Article Using IEEEtran.cls for IEEE Journals}


\maketitle

\begin{abstract}
The success of existing salient object detection models relies on a large pixel-wise labeled training dataset, which is time-consuming and expensive to obtain.
We study semi-supervised salient object detection, with access to a small number of labeled samples and a large number of unlabeled samples. 
Specifically, we present a pseudo label based learning framework with a Conditional Energy-based Model. We model the stochastic nature of human saliency labels using the stochastic latent variable of the Conditional Energy-based Model. It further enables generation of a high-quality pixel-wise uncertainty map, highlighting the reliability of corresponding pseudo label generated for the unlabeled sample. This minimises the contribution of low-certainty pseudo labels in optimising the model, preventing the error propagation.
Experimental results show that the proposed strategy can effectively explore the contribution of unlabeled data. With only 1/16 labeled samples,
our model achieves competitive performance compared with state-of-the-art fully-supervised models.
\end{abstract}



\section{Introduction}
\label{sec:intro}
\IEEEPARstart{A}{s} a class-agnostic image segmentation task, salient object detection (SOD) \cite{borji2015salient,mei2021exploring,hu2020sac,sun2021ampnet} aims to localise the full scope of objects that attract human attention. Most existing SOD models are fully-supervised \cite{zhang2018bi,feng2019attentive,wang2019salient}, where a pixel-wise labeled dataset \cite{DUTS-TE} is used to learn a mapping function from the input space to the output space. Within this setting, methods usually focus on effective feature aggregation \cite{zhang2017supervision,zhang2018bi,wang2018detect,wu2019mutual} for structure-preserving saliency prediction \cite{feng2019attentive,wang2019salient}.
The success of the fully-supervised methods comes from large pixel-wise labeled training datasets \cite{wang2018detect,UC-Net,wu2019mutual,zhang2018bi}, which are both time-consuming and expensive to obtain. To relieve the annotation burden, some research focuses on the weakly-supervised saliency detection with easy-to-obtain annotations, e.g.,~image level labels \cite{ASMO,DUTS-TE}, scribbles \cite{Scribble_Saliency}, etc. In this paper, we consider the semi-supervised learning approach to easing \cite{FC-SOD,SAL} the labeling burden, which aims to train a model
with only a small number
of labelled samples and a large number
of unlabeled samples.

\begin{figure}[!t]
\scriptsize
\centering
    \begin{tabular}{c@{ }c@{ }c@{ }c@{ }}
        \includegraphics[width=0.235\linewidth]{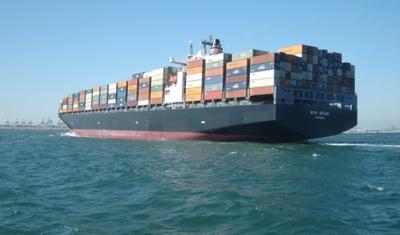}&
        \includegraphics[width=0.235\linewidth]{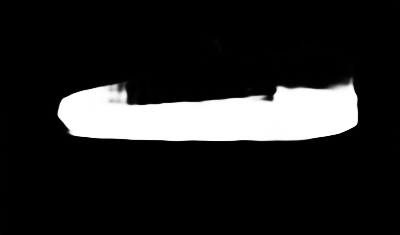}&
        \includegraphics[width=0.235\linewidth]{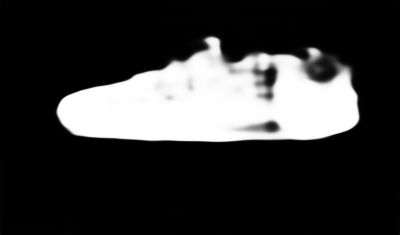}&
        \includegraphics[width=0.235\linewidth]{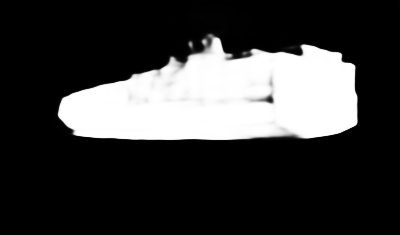}\\
        \includegraphics[width=0.235\linewidth]{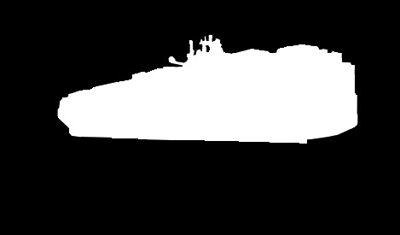}&
        \includegraphics[width=0.235\linewidth]{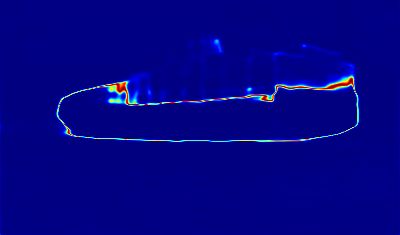}&
        \includegraphics[width=0.235\linewidth]{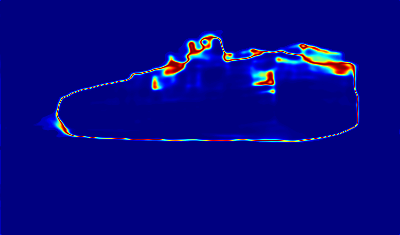}&
        \includegraphics[width=0.235\linewidth]{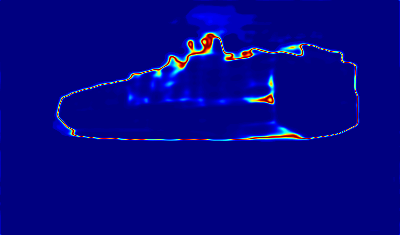}\\
        \scriptsize{Img / GT} & \scriptsize{VAE} & \scriptsize{GAN} & \scriptsize{Ours}\\
    \end{tabular}
    \caption{
        Pseudo-labels and uncertainty map of the unlabeled training sample estimated by three generative models including VAE \cite{auto_encoding_variational_bayes}, GAN \cite{GAN} and our
        approach (Ours).
        The isotropic Gaussian distribution of the latent variable within both VAE and GAN leads to less accurate uncertainty maps. Our model generates the highest-quality pseudo label with a correct uncertainty map highlighting the inaccurate regions.
    }
\label{fig:introduction_visualisation}
\end{figure}


Existing Semi-Supervised Learning (SSL) methods can be roughly categorised into (i) consistency regularisation based \cite{S4L,noisy_student,EnAET}, and (ii) pseudo label based \cite{FC-SOD,SSGM,ECS}. Consistency regularisation based approaches design a rich set of data augmentation techniques and enforce consistent predictions across a range of augmented samples. Its success relies on designing a rich set of domain-specific data augmentation techniques \cite{CCT,DCC,CPS}. This entails domain-specific prior knowledge and often has limited generalisation beyond the source domain. On the other hand, pseudo-label based approaches \cite{FC-SOD,SSGM,ECS} do not require domain specific augmentations, making them easier to implement. However, pseudo labels can be can include incorrect labels, often regarded as noise, at various scales. Directly incorporating noisy pseudo labels into the model optimisation causes error propagation, resulting in a biased model that over-fits to label noise \cite{ATSO}. To achieve an effective pseudo-label based semi-supervised learning, the model should be aware of the uncertainties of the corresponding pseudo labels.

The earliest Semi-Supervised Salient Object Detection (SS-SOD) work \cite{SaliencyGAN} proposes a framework composed of two sets of General Adversarial Networks \cite{GAN} (GAN). The first GAN tasks its generator to synthesise more training samples and its discriminator to discriminate between real and synthesised training data. In the second GAN, the generator is trained for saliency prediction and the discriminator distinguishes between the predictions for labeled and unlabeled data. Subsequent SS-SOD works are pseudo-label based. FC-SOD \cite{FC-SOD} presents an Adversarial-Paced Learning (APL) framework, consisting of a task predictor and a pace generator. The former is tasked to learn from labeled data to generate pseudo-labels for unlabeled data. The latter, optimised with adversarial training, predicts a reliability map for each unlabeled data, which in turn serves as the criteria to select reliable pseudo labels to update the model.
SAL \cite{SAL} develops an active learning framework. Unlabeled data that are closer to labeled data in feature space are progressively included to optimise the model. It also employs an auxiliary network to predict an uncertainty map through an adversarial learning.

A common drawback in both pseudo-label based SS-SOD methods lies in the quality of the reliability/uncertainty map. The reliability/uncertainty map, obtained with an auxiliary network through adversarial learning, is downsampled, not matching the resolution of the training image. The resultant reliability/uncertainty map lacks fine-grained details that are important to a dense classification task. On the other hand, \cite{Liu_2022_WACV} has pointed out that learning an auxiliary network with a U-Net structure for uncertainty estimation via adversarial learning results in a trial solution and incorrect uncertainty map. To address this, we propose a Conditional Energy Based Model (CEBM), where we use the stochastic latent variable to model the naturally stochasticity of the human labelling process particularly from multiple annotators, and further exploit it to compute the epistemic uncertainty with fine-grained details. The resultant high-resolution uncertainty map allows us to achieve a more effective utilisation of unlabeled samples by reducing the scale of error propagation \cite{ATSO}.

We summarise our main contributions as: (1) We propose a pseudo-labeled based SS-SOD learning framework with a Conditional Energy-Based Model (CEBM); (2) We use the stochastic latent variable to model the natural stochasticity of the human labelling process particularly from multiple annotators, and further compute the epistemic uncertainty on the pseudo labels with fine-grained details; (3) We propose a confidence-aware semi-supervised learning on the unlabeled data with pseudo labels, preventing the error propagation;
(4) We demonstrate that the proposed solution achieves state-of-the-art performance on the SS-SOD task.



\section{Related Work}
\subsection{Salient Object Detection}
Early SOD approaches \cite{wang2015deep,li2015visual,zhao2015saliency} adopted multi-layer perceptrons (MLPs) to predict the saliency maps. They are generally outperformed by more recent fully convolutional network (FCN) based methods which leverage the more powerful representation ability of CNNs. Feature aggregation \cite{zhang2017supervision,zhang2018bi,wang2018detect,wu2019mutual,zhou2021ecffnet,li2021uncertainty} and deep supervision \cite{amulet,hou2017deeply,hu2018recurrently,islam2018revisiting,wang2019quality} are commonly adopted in these methods for accurate prediction generation. There are also works that incorporate additional information to preserve structures of the salient objects, e.g., object boundary \cite{feng2019attentive,wang2019salient,jiang2020cmsalgan}, fixation \cite{wang2018salient,kruthiventi2016saliency}. Another branch of research \cite{PicaNet,zhao2019pyramid} uses the attention mechanism to exploit the local-global dependencies in handling the SOD task, i.e., vision transformer \cite{zhang2021learning_nips,mao2021transformer,liu_ICCV_2021_VST,liu2021swinnet}.

\subsection{Semi-Supervised Learning}
Semi-Supervised Learning (SSL) addresses the annotation shortage by incorporating a large amount of unlabeled data in training a classifier. It has been extensively studied for classification tasks with many works focusing on consistency based \cite{ladder_network,pi_model,temporal_ensemble,mean_teacher,virtual_adversarial_training,unsupervised_data_augmentation,CCT,DCC,CPS,S4GAN,ATSO} and pseudo-label based \cite{co-training,pseudo_label,noisy_student,S4L,EnAET} techniques. Both of these methods initiate the training on the labeled data and differ
in terms of the usage of the unlabeled part. Consistency based methods constrain the model to produce similar predictions for unlabeled training samples under different data augmentation schemes.
Pseudo-label based methods usually perform semi-supervised learning by exploring the contribution of unlabeled samples for model updating via generating reliable pseudo labels for them.
Although semi-supervised learning has been well-explored for image classification, it's still under-explored for salient object detection.
SaliencyGAN \cite{SaliencyGAN} designs a semi-supervised SOD model via adversarial training, where an auxiliary image generator is used to generate a meaningful image representation, with which the unlabeled samples can participate into model updating via the adversarial loss term.
Similarly, FC-SOD~ \cite{FC-SOD} proposes an adversarial-paced learning scheme that gradually explore the contribution of the unlabeled set for semi-supervised SOD.
SAL \cite{SAL} combines semi-supervised learning with active learning for progressive semi-supervised SOD, where the samples that cannot be well represented by the labeled samples will be labeled, leading to a gradually enlarged labeled pool.

\subsection{Deep Generative Model}
Due to the one-to-distribution mapping attribute, deep generative models~\cite{GAN,auto_encoding_variational_bayes,sohn2015learning,LeCun06atutorial,normalizing_flow,ddpm,han2017alternating} have been explored in many computer vision tasks to produce stochastic predictions, and the success has also been extended to salient object detection~\cite{li2019supervae,UC-Net} for effective background reconstruction~\cite{li2019supervae} or saliency subjective nature modeling~\cite{UC-Net}.
For semi-supervised segmentation, 
 GANs have also been used in \cite{gan_semi_seg,semanticGAN}. VAEs have been explored in \cite{vae_amodal_obj_completion} for amodal object completion without access to the amodal labels at training time. For both types of generative models based semi-supervised frameworks, the basic assumption is that the latent variables that model the data distribution should follow an isotropic Gaussian distribution, which may be biased, leading to less informative latent space.

\subsection{Uncertainty Estimation}
Uncertainty estimation has been studied to represent the trustworthiness of model predictions \cite{kendall2017uncertainties,guo2017calibration}. A main approach involves generating multiple predictions and computes the variance across them as the uncertainty. Monte Carlo Dropout (MC-Dropout) \cite{kendall2017uncertainties,Durasov21_maskensemble} produces a set of distinct predictions by randomly sampling network connections to drop. On the other hand, ensemble-based methods \cite{lakshminarayanan2017simple,snapshot_ensembles,Wen2020BatchEnsemble} achieves multiple predictions with an ensemble of models. Recent research finds that MC-Dropout usually fails to explore the true representation of the model uncertainty, and the performance of ensemble-based methods are highly dependent on the design of the ensemble structure. The latent variable model \cite{auto_encoding_variational_bayes,GAN,han2017alternating} presents an alternative to the aforementioned methods to estimate the uncertainty, where distinguishing predictions are enabled by the stochastic nature of latent variable.

Uncertainty estimation has also been applied in the dense classification tasks. \cite{li2021uncertainty,FC-SOD,SAL} uses an auxiliary network with an adversarial training to estimate the uncertainty. OCENet\cite{Liu_2022_WACV} designs a network to model the distributional gap between the predictions and groundtruth labels as uncertainty. \cite{Kwon_2022_CVPR} also approximates the prediction-to-groundtruth gap in a semi-supervised semantic segmentation framework. On the other front, \cite{GCoSOD} proposes a self-supervised method, associating diverse groundtruth labels with each training sample, to estimate prediction uncertainty.


\begin{figure*}[t!]
\centering
\includegraphics[width=\textwidth]{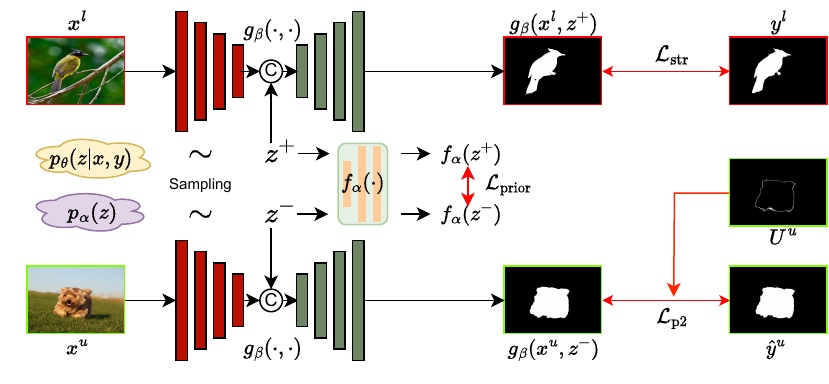}
\caption{
The proposed CEBM framework for SS-SOD is composed of a Saliency Prediction Model ($g_{\beta}(\cdot, \cdot)$) and a Prior Model ($f_{\alpha}(\cdot)$). Its training is divided into two phases: (1) optimisation on the labeled samples, $(x^{l}, y^{l})$, updates: (a) the Prior Model with $\mathcal{L}_{\text{prior}}$ (Eq.~\ref{eq:energy_difference_loss}), allowing it align the prior distribution ($p_{\alpha}(z)$) towards to posterior distribution ($p_{\theta}(z | x, y)$) which approximates the distribution of the labeled samples ($p(x^{l}, y^{l})$) in the latent space, and (b) the Saliency Prediction Model with $\mathcal{L}_{\text{str}}$ (Eq.~\ref{eq:structure_loss}), allowing it to decode the posterior latent variable ($z^{+}$) into a saliency prediction conditioned on the input image; (2) confidence-aware learning on the unlabeled samples with pseudo labels, $(x^{u}, \hat{y}^{u})$, 
updating only the Saliency Prediction Model with $\mathcal{L}_{\text{p2}}$ (Eq.~\ref{eq:phase_two_loss}). Only prior latent variable $z^{-}$ is sampled to produce saliency prediction on the unlabeled samples ($g_{\beta}(x^{u}, z^{-})$) and $U^{u}$ denotes the uncertainty map of the corresponding pseudo label.
}
\label{fig:Learning_Framework}
\end{figure*}

\section{Our Method}
We first introduce the task setting in Sec.~\ref{sub_sec:task_setting}, and the optimisation of the Conditional Energy-Based Model in Sec.~\ref{sub_sec:optimisation_of_CEBM}, which is mainly adapted from the optimisation of EBM model in \cite{pang2020learning}. The main contribution is an Conditional Energy-Based Model framework for SS-SOD as delineated in Sec.~\ref{sub_sec:framework}, Sec.~\ref{sec:labeled_data_distribution_approximation} and Sec.~\ref{sec:confidence_aware_semi_supervised_learning}. We also include the network details in Sec.~\ref{sub_sec:model_details}.

\subsection{Task Settings}
\label{sub_sec:task_setting}
For semi-supervised salient object detection, we have a dataset $D = \{D^{l}, D^{u}\}$ consisting of $N_l$ labeled images $D^l=\{x_i^l,y_i^l\}_{i=1}^{N_l}$, and $N_u$ unlabeled images $D^u=\{x_j^u\}_{j=1}^{N_u}$, where $x_{i}^{l} \in \mathbbm{R}^{3 \times H \times W}, y_{i}^{l} \in \{0, 1\}^{H \times W}$ are a pair of a labeled image and its corresponding ground truth saliency map, and $x_{j}^{u} \in \mathbbm{R}^{H \times W}$ represents the unlabeled image. Typically, the numbers of labeled samples is significantly less than the number of unlabeled samples $N_{l} \ll N_{u}$. 
Our proposed CEBM is composed of a Saliency Prediction Model and a Prior Model, parameterised by $\beta$ and $\alpha$ respectively. The Prior Model takes a latent variable, $z$, as input and outputs an energy score, $f_{\alpha}(z): \mathbbm{R}^{d} \rightarrow \mathbbm{R}^{+}$, where $d$ is the dimension of the latent space. The Saliency Prediction Model takes an input image and a latent variable to predict the saliency map, $g_{\beta}(x, z): \mathbbm{R}^{3 \times H \times W} \rightarrow (0, 1)^{H \times W}$. We use $\theta = (\alpha, \beta)$ to denote the overall model parameters.

\subsection{Preliminary on Energy-Based Model}
\label{sub_sec:optimisation_of_CEBM}
This section introduces our sampling strategy from the latent space of Energy-Based Model (EBM) for the SS-SOD task and necessary preliminaries on the optimisation of EBM \cite{pang2020learning}. Let $z \in \mathbbm{R}^{d}$ be a latent variable, and $p_{\alpha}(z)$ and $p_{\theta}(z | x, y)$ be its prior distribution and posterior distribution respectively, where the later approximates the joint distribution of training images and their corresponding groundtruth saliency maps in the latent space. 

During the training on labeled samples, we sample from both the prior and the posterior latent distributions and align the prior distribution towards the posterior distribution. This is achieved by optimising the Prior Model with the energy difference between the prior latent variable and the posterior latent variable \cite{pang2020learning,zhang2021learning_nips} as:
\begin{equation}
\delta_{\alpha}(x) = \mathbb{E}_{p_{\theta}(z | x, y)} \Bigl[ \nabla_{\alpha} f_{\alpha}(z) \Bigr] - \mathbb{E}_{p_{\alpha}(z)} \Bigl[ \nabla_{\alpha} f_{\alpha}(z) \Bigr].
\label{eq:learning_gradient_for_prior_model}
\end{equation}
Both expectations over the prior and posterior distributions of the latent variable are intractable. We follow \cite{pang2020learning,zhang2021learning_nips} to approximate these expectations via Markov-Chain Monte-Carlo (MCMC) sampling \cite{nijkamp2019learning,nijkamp2020learning} of the latent variables from the respective distributions. To sample from the prior distribution $p_{\alpha}(z)$ and the posterior distribution $p_{\theta}(z | x, y)$, it starts from a latent variable from a Gaussian distribution, $z_{0} \sim p_{0}(z) = \mathcal{N}(0, \sigma^{2}I_{d})$, and then update it with Langevin dynamics \cite{zhu1998grade,neal2011mcmc} as:
\begin{equation}
\begin{aligned}
& z_{k + 1}^{-} = z_{k}^{-} + s \nabla_{z} \log p_{\alpha} (z_{k}^{-}) + \sqrt{2s} \epsilon_{k},\\
& z_{k + 1}^{+} = z_{k}^{+} + s \nabla_{z} \log p_{\theta}(z_{k}^{+} | x, y) + \sqrt{2s} \epsilon_{k},\\
& \qquad k = 1, \dots, K, \qquad \epsilon_{k} \sim \mathcal{N}(0, I_{d}),
\label{eq:Langevin_dynamics_update}
\end{aligned}
\end{equation}
where $z^{-}$ and $z^{+}$ denote the prior and posterior latent variable respectively and both of them start from an initial latent variable $z_{k = 0}^{-} = z_{k = 0}^{+} = z_{0}$, $K$ is the total number of Langevin dynamics steps, $s$ is the Langevin dynamics step size. Both $\nabla_{z} \log p_{\alpha} (z_{k}^{-})$ and $\nabla_{z} \log p_{\theta} (z_{k}^{+} | x, y)$ can be computed with back-propagation in implementation \cite{pang2020learning,zhang2021learning_nips}.

Beside the training on labeled samples, the latent variable is sampled only from its prior distribution, which is assumed to have already been aligned with its posterior distribution after being optimised on the labeled data \cite{pang2020learning}. This includes the (i) generation of pseudo labels on the unlabeled samples; (ii) prediction during training on unlabeled samples; (iii) inference on testing data. In both (i) and (iii), we do not have the labels that are necessary to construct the posterior distribution, whereas in (ii), we avoid aligning the prior distribution towards a posterior distribution based on the noisy pseudo labels. That means the Prior Model is not optimised while training on the unlabeled samples with pseudo labels. On the other hand, we sample multiple latent variables for each input image in (i) to produce equally many predictions and adopt their mean as the pseudo label for the unlabeled samples. For both (ii) and (iii), we sample a single prior latent variable to generate a single saliency prediction for each input image. 

The Saliency Prediction Model is optimised on both labeled and unlabeled data, with groundtruth labels and pseudo labels respectively. Following \cite{pang2020learning,zhang2021learning_nips}, the learning gradient for the Saliency Prediction Model can be computed as:
\begin{equation}
\delta_{\beta}(y | x) = \mathbb{E}_{p_{\theta}(z | x, y)} \Bigl[ \nabla_{\beta} \log p_{\beta}(y | x, z) \Bigr],
\label{eq:learning_gradient_of_saliency_prediction_model}
\end{equation}
where $\log p_{\beta}(y | x, z)$ is the objective function.

\subsection{Framework}
\label{sub_sec:framework}
As shown in Fig.~\ref{fig:Learning_Framework}, we introduce a Conditional Energy-Based Model (CEBM) for SS-SOD. The CEBM is composed of two components: (1) a U-Net based Saliency Prediction Model $g_{\beta}(x, z): \mathbbm{R}^{3 \times H \times W} \rightarrow (0, 1)^{H \times W}$ parameterised by $\beta$, and (2) a Conditional Energy-Based Prior Model (abbreviated as Prior Model) $f_{\alpha}(z): \mathbbm{R}^{d} \rightarrow \mathbbm{R}$ parameterised by $\alpha$, where $d$ is the dimension of the latent variable. The proposed CEBM extends the Energy Based Model presented in \cite{pang2020learning} to produce a saliency map from a latent variable $z$, conditioned on the input image $x$. 
The training pipeline is divided into two stages: (1) optimise the CEBM model with only labeled data with fully-supervised learning (see Sec.~\ref{sec:labeled_data_distribution_approximation}), and then (2) employ the proposed Confidence-aware Semi-Supervised Learning (Sec.~\ref{sec:confidence_aware_semi_supervised_learning}) to optimise the CEBM model with unlabeled data.

\subsection{Model the Distribution of Labeled Data}
\label{sec:labeled_data_distribution_approximation}
In the first phase of training, the CEBM model is optimised with only the labeled data in a fully supervised setting, allowing the Prior Model to approximate the distribution of the labeled samples. We first sample a single latent variable from the initial latent variable distributions $z_{0}^{- / +} \sim \mathcal{N}(0, I_{d})$. Then it can be iteratively updated with Langevin dynamics defined in Eq.~\ref{eq:Langevin_dynamics_update} towards the prior distribution ($z^{-}_{K} \sim p_{\alpha}(z)$) and the posterior distribution $z^{+}_{K} \sim p_{\theta}(z | x, y)$ respectively. The posterior latent variable is employed in predicting the saliency map $g_{\beta}(x, z_{K}^{+})$ for the input image $x$.

The Saliency Prediction Model is optimised with a Structure Loss, which is a combination of a Binary Cross Entropy (BCE) loss and a Dice loss \cite{dice_loss}, as:
\begin{equation}
    \mathcal{L}_{\text{str}} = \mathcal{L}_{\text{bce}} \bigl(g_{\beta}(x^{l}, z_{k}^{+}), y^{l} \bigr) + \mathcal{L}_{\text{dice}} \bigl(g_{\beta}(x^{l}, z_{k}^{+}), y^{l} \bigr),
\label{eq:structure_loss}
\end{equation}
where BCE loss is defined as
\begin{equation}
\begin{aligned}
\mathcal{L}_{\text{bce}} (a, b) = \frac{1}{HW} \sum_{i = 1, j = 1}^{H, W} & \Bigl( - b_{i, j} \log(a_{i, j})\\
& - (1 - b_{i, j}) \log(1 - a_{i, j}) \Bigr), \\
\end{aligned}
\label{eq:BCE_loss}
\end{equation}
and Dice loss is defined as:
\begin{equation}
\begin{aligned}
\mathcal{L}_{\text{dice}} (a, b) = 1 - \frac{2 \cdot \frac{1}{HW} \sum_{i = 1, j = 1}^{H, W} \bigl( b_{i, j} \cdot a_{i, j} \bigr) + \epsilon}{\frac{1}{HW} \sum_{i = 1, j = 1}^{H, W} \bigl( b_{i, j} + a_{i, j} \bigr) + \epsilon},
\end{aligned}
\label{eq:dice_loss}
\end{equation}
$\epsilon$ is a small number, that is empirically set to $1e-8$, to avoid division by zero.

The prior model is optimised with the energy energy difference between the posterior latent variable and the prior latent variable. Its objective function is defined as:
\begin{equation}
\mathcal{L}_{prior}(z^{-}, z^{+}) = f_{\alpha}(z^{+}) - f_{\alpha}(z^{-}).
\label{eq:energy_difference_loss}
\end{equation}
Intuitively, Eq.~\ref{eq:energy_difference_loss} aligns the prior distribution of latent variable towards the posterior distribution of latent variable, which approximates the joint distribution of labeled samples $p(X, Y)$ in the latent space.

\subsection{Confidence-Aware Semi-Supervised Learning}
\label{sec:confidence_aware_semi_supervised_learning}

The Confidence-Aware Semi-Supervised Learning (CA-SSL) relies on the stochastic attribute of the Saliency Prediction Model.
The stochasticity lies in the noise introduction in the Langevin Dynamics updates and the sampling of the latent variable. This models the stochasticity of human saliency, particularly with the different human annotators. Unlike the supervised learning on the labeled samples, we sample only the prior latent variable from the prior distribution, $z^{-} \sim p_{\alpha}(z)$, to generate the pseudo label on the unlabeled samples. We assume that the prior distribution, having been updated during the supervised learning on the labeled samples, can approximate the joint distribution of labeled samples, and do not want to tailor it towards a noisy joint distribution of the unlabeled samples.

To generate the pseudo labels on the unlabeled samples, we first sample $M$ 
prior latent variables, $z_{l}^{-} \sim p_{\alpha}(z), l = 1, \dots, M$ (here we omit the Langevin dynamics step index $k$ and use $l$ to index the number of prior latent variables), to obtain 
$M$
predicted saliency maps whose mean are taken as pseudo labels of the unlabeled data:
\begin{equation}
    \hat{y}^{u} = \frac{1}{M} \sum_{m=1}^{M} g_{\beta}(x^{u}, z_{m}^{-}), \quad z_{m} \sim p_{\alpha}(z)
    \label{eq:pseudo_label_generation}
\end{equation}
where we empirically set $M = 10$. We adopt the soft pseudo labels, which encode richer information, to further optimise the CEBM model. 
The epistemic uncertainty \cite{kendall2017uncertainties} is represented as variance in model predictions, which, after being averaged, yields prediction with high entropy values. Thus, we further compute the entropy of the pseudo label as uncertainty:
\begin{equation}
    U^{u} = \mathbb{H} \bigl( \hat{y}^{u} \bigr) = - \hat{y}^{u} \log_{2}(\hat{y}^{u}) - (1 - \hat{y}) \log_{2} (1 - \hat{y}^{u}) 
    \label{eq:prediction_uncertainty}
\end{equation}
where $\mathbb{H}(\cdot)$ is an entropy function. The resultant uncertainty is in the range $U^{u} \in (0, 1)^{H \times W}$. The prediction confidence can be obtained by inverting the prediction uncertainty:
\begin{equation}
    C^{u} = \mathbbm{J}_{H, W} - U^{u}
    \label{eq:prediction_confidence}
\end{equation}
where $\mathbbm{J}_{H, W}$ is an all-one matrix with $H$ rows and $W$ columns. Higher prediction uncertainty corresponds to lower prediction confidence and vice versa.

We apply a confidence-aware loss 
in the second phase of optimising the CEBM model with unlabeled samples:
\begin{equation}
    \mathcal{L}_{\text{str-c}} = C^{u} \cdot \Bigl(  \mathcal{L}_{\text{bce}} \bigl(g_{\beta}(x^{u}, z_{k}^{-}), y^{u} \bigr) + \mathcal{L}_{\text{dice}} \bigl(g_{\beta}(x^{u}, z_{k}^{-}), y^{u} \bigr) \Bigr),
\end{equation}
where the latent variable is sampled from the prior distribution in the absence of the groundtruth label. The confidence is applied pixel-wise to screen the noisy parts of pseudo labels from participating in updating the models. An inherent issue with learning from soft pseudo labels is that the resultant saliency prediction is overly smoothed. Despite the correctness of these predictions, they often lead to large mean absolute error. As a solution to this, we further incorporate an entropy loss to encourage the Generator model to produce more confident predictions. Overall, the learning objective of the second training phase is:
\begin{equation}
    \mathcal{L}_{\text{p}_{2}} = \lambda_{\text{us}} \cdot \mathcal{L}_{\text{str-c}} + \lambda_{\text{ue}} \cdot \mathbb{H} \bigl( g_{\beta}(x^{u}, z_{k}^{-}) \bigr)
\label{eq:phase_two_loss}
\end{equation}
where $\lambda_{\text{us}}$ and $\lambda_{\text{ue}}$ are hyperparameters balancing the two components and we empirically set both of them to 1.

The Prior Model is not optimised when training on the unlabeled samples with pseudo labels. This is because the prior distribution cannot be effectively updated to approximate the joint distribution of unlabeled images where the groundtruth labels are missing.

\subsection{Model Details}
\label{sub_sec:model_details}
\noindent\textbf{Saliency Prediction Model}: 
We build the Saliency Prediction Model $g_\beta(x,z)$ with a ResNet50 backbone,
which can produce four levels of backbone features: $\{s_k\}_{k=1}^4$. We adopt the decoder of MiDaS \cite{MiDaS}, which consists of four fusion blocks ($\{F_{i}\}_{i=1}^{4}$). The first fusion block upscales the highest-level feature ($\bar{s_{4}} = F_{4}(s_{4})$) by a factor of two. The subsequent fusion blocks gradually fuse feature pairs and upscale the resultant feature by a factor of two to produce $\{(\bar{s}_{k-1} = F_{k-1}(\bar{s}_{k}, s_{k-1}))\}_{k=4}^{2}$. The output layer adopts the final feature $\bar{s}_{1}$ and produces a binary saliency map $y'$ of the same spatial size as the input image. 

\noindent\textbf{Prior Model}: 
The Prior Model  ($p_{\alpha}$) is parameterised by a multi-layer perceptron (MLP) to model the energy function in latent space. It takes the latent variable as input and outputs an energy value. The model consists of three fully-connected layers of dimensions $\{d_{z}, 100, 100\}$ where $d_{z}$ is the dimension of the latent variable. Each fully connected layer, except the last one, is followed by a GELU activation function \cite{gelu}.

\section{Experimental Results}
\noindent\textbf{Dataset:}
We train our proposed semi-supervised learning method on the DUTS \cite{DUTS-TE} training dataset with six different labeled set split ratios, including 1/64, 1/32, 1/16, 1/8, 1/4 and 1/2, following the typical settings \cite{pseudoseg,AEL,zhou2021c3}. Our method is further evaluated on six SOD benchmark datasets, including the DUTS test set \cite{DUTS-TE}, DUT-OMRON \cite{DUT-OMRON}, PASCAL-S \cite{PASCAL-S}, SOD \cite{SOD}, ECSSD \cite{shi2015hierarchical} and HKU-IS \cite{li2015visual}. We report our benchmark performance trained using a labeled ratio of 1/16 ($N_l=659$) in Tab.~\ref{tab:SS-SOD_benchmark}.

\noindent\textbf{Evaluation metrics:}
Following previous works,
\cite{FC-SOD, SaliencyGAN}, 
we adopt maximum F-measure ($F_{\xi}^{\mathrm{max}}$) and mean absolute error (MAE $\mathcal{M}$) as our evaluation metrics.

\noindent\textbf{Implementation details:}
1) Training details: We implement our method in the Pytorch framework. The encoder (ResNet-50 \cite{resnet} backbone) is initialised with
weights pretrained on ImageNet. The decoder and the energy-based model are initialised by default. Our proposed method is trained with the Adam optimiser \cite{kingma2014adam}. All images are rescaled to $480 \times 480$ in resolution. For both phase (1) and phase (2) training, the learning rates of the generator model and the energy-based model are initialised to $l_{G} = 2.5 \times 10^{-5}$ and $l_{E} = 1 \times 10^{-5}$ respectively. The learning rates are decayed by a factor of 0.9 after every 1,000 iterations. With a fixed batch size of 8, the training is composed of 6,500 iterations in phase (1) and 8,500 iterations in phase (2). The complete training takes 7 hours using a single NVIDIA Geforce RTX 3090 GPU.
2) Semi-supervised model related hyper-parameters: In our experiment, we run $K^-=K^+=5$ Langevin steps with step size $\delta^-=0.4$ and $\delta^+=0.1$ for the prior and posterior model respectively. Empirically, we set the hyperparameter $\sigma^2_z=1.0$ and $\sigma^2_\epsilon=0.3$.

\begin{table*}[t!]
  \centering
  \scriptsize
  \renewcommand{\arraystretch}{1.3}
  \renewcommand{\tabcolsep}{1.3mm}
  \caption{Performance comparison with state-of-the-art fully-/weakly-/semi-supervised SOD methods on benchmark testing dataset. P - pixel-level label; I - image-level label; C - caption; S - scribble. Our model achieves significant improvements over the semi-supervised models, especially on F-max
  . Comparisons with FC-SOD on the same data split ratios are shown in Fig.~\ref{fig:ablation_study_different_dataset_splits}.}
  \begin{tabular}{clccccccccccccc}
  \toprule
  \multicolumn{2}{c}{\multirow{2}{*}{Methods}} & 
  \multirow{2}{*}{Label}& \multicolumn{2}{c}{DUTS-TE} & \multicolumn{2}{c}{DUT-OMRON} & \multicolumn{2}{c}{PASCAL-S} & \multicolumn{2}{c}{SOD} & \multicolumn{2}{c}{ECSSD} & \multicolumn{2}{c}{HKU-IS}\\
  & & & $F_{\xi}^{\mathrm{max}}\uparrow$ & $\mathcal{M}\downarrow$ & $F_{\xi}^{\mathrm{max}}\uparrow$ & $\mathcal{M}\downarrow$ & $F_{\xi}^{\mathrm{max}}\uparrow$ & $\mathcal{M}\downarrow$ & $F_{\xi}^{\mathrm{max}}\uparrow$ & $\mathcal{M}\downarrow$ & $F_{\xi}^{\mathrm{max}}\uparrow$ & $\mathcal{M}\downarrow$ & $F_{\xi}^{\mathrm{max}}\uparrow$ & $\mathcal{M}\downarrow$ \\
  \midrule
  \multirow{6}{*}{\parbox{1.2cm}{Fully-Supervised}}
  & CPD \cite{CPD} & P & 0.865 & 0.043 & 0.797 & 0.056 & 0.859 & 0.071 & 0.860 & 0.112 & 0.939 & 0.037 & 0.925 & 0.034 \\ 
  & BASNet \cite{BASNet} & P & 0.859 & 0.048 & 0.805 & 0.056 & 0.854 & 0.076 & 0.851 & 0.114 & 0.942 & 0.037 & 0.928 & 0.032\\
  & PoolNet \cite{PoolNet} & P & 0.888 & 0.039 & 0.815 & 0.053 & 0.863 & 0.075 & - & - & 0.944 & 0.039 & 0.932 & 0.033\\
  & EGNet \cite{EGNet} & P & 0.888 & 0.039 & 0.815 & 0.053 & 0.865 & 0.074 & 0.889 & 0.099 & 0.947 & 0.037 & 0.935 & 0.031\\
  & MINet \cite{MINet} & P & 0.884 & 0.037 & 0.810 & 0.056 & 0.867 & 0.064 & - & - & 0.947 & 0.033 & 0.935 & 0.029\\
  & ITSD \cite{ITSD} & P & 0.882 & 0.041 & 0.820 & 0.061 & 0.870 & 0.064 & - & - & 0.947 & 0.034 & 0.934 & 0.031\\
  \hline
  \multirow{4}{*}{\parbox{1.2cm}{Weakly-Supervised}} 
  & WSS \cite{DUTS-TE} & I & 0.740 & 0.099 & 0.695 & 0.110 & 0.773 & 0.140 & 0.778 & 0.171 & - & - & - & -\\
  & MWS \cite{MWS} & I \& C & 0.768 & 0.091 & 0.722 & 0.108 & 0.786 & 0.134 & 0.801 & 0.170 & 0.878 & 0.096 & 0.856 & 0.084\\
  & ASMO \cite{ASMO} & I & - & - & 0.732 & 0.100 & 0.758 & 0.154 & 0.758 & 0.187 & - & - & - & -\\
  & SS \cite{Scribble_Saliency} & S & 0.789 & 0.062 & 0.753 & 0.068 & 0.811 & 0.092 & 0.806 & 0.131 & 0.888 & 0.061 & 0.880 & 0.047\\
  \hline
  \multirow{3}{*}{\parbox{1.2cm}{Semi-Supervised}} 
  & SaliencyGAN \cite{SaliencyGAN} & 1,000 & 0.641 & 0.135 & 0.610 & 0.131 & 0.699 & 0.164 & - & - & 0.776 & 0.116 & 0.742 & 0.107 \\
  & SAL \cite{SAL} & 1,500 & 0.849 & 0.046 & 0.766 & \textbf{0.062} & 0.804 & 0.088 & 0.835 & 0.105 & 0.884 & 0.057 & 0.920 & \textbf{0.032}\\
  & FC-SOD \cite{FC-SOD} & 1,000 & 0.846 & 0.045 & 0.767 & 0.067 & 0.848 & 0.067 & 0.846 & 0.122 & 0.914 & 0.047 & 0.903 & 0.038 \\
  \hline
  \multirow{1}{*}{Ours} & CEBM & \textbf{659 ($\frac{1}{16}$)} & \textbf{0.869} & \textbf{0.045} & \textbf{0.774} & 0.067 & \textbf{0.864} & \textbf{0.067} & \textbf{0.856} & \textbf{0.104} & \textbf{0.931} & \textbf{0.043} & \textbf{0.928} & 0.034 \\
  \bottomrule
  \end{tabular}
  \label{tab:SS-SOD_benchmark}
\end{table*}

\begin{figure*}[ht!]
    \centering
    \includegraphics[width=0.95\textwidth]{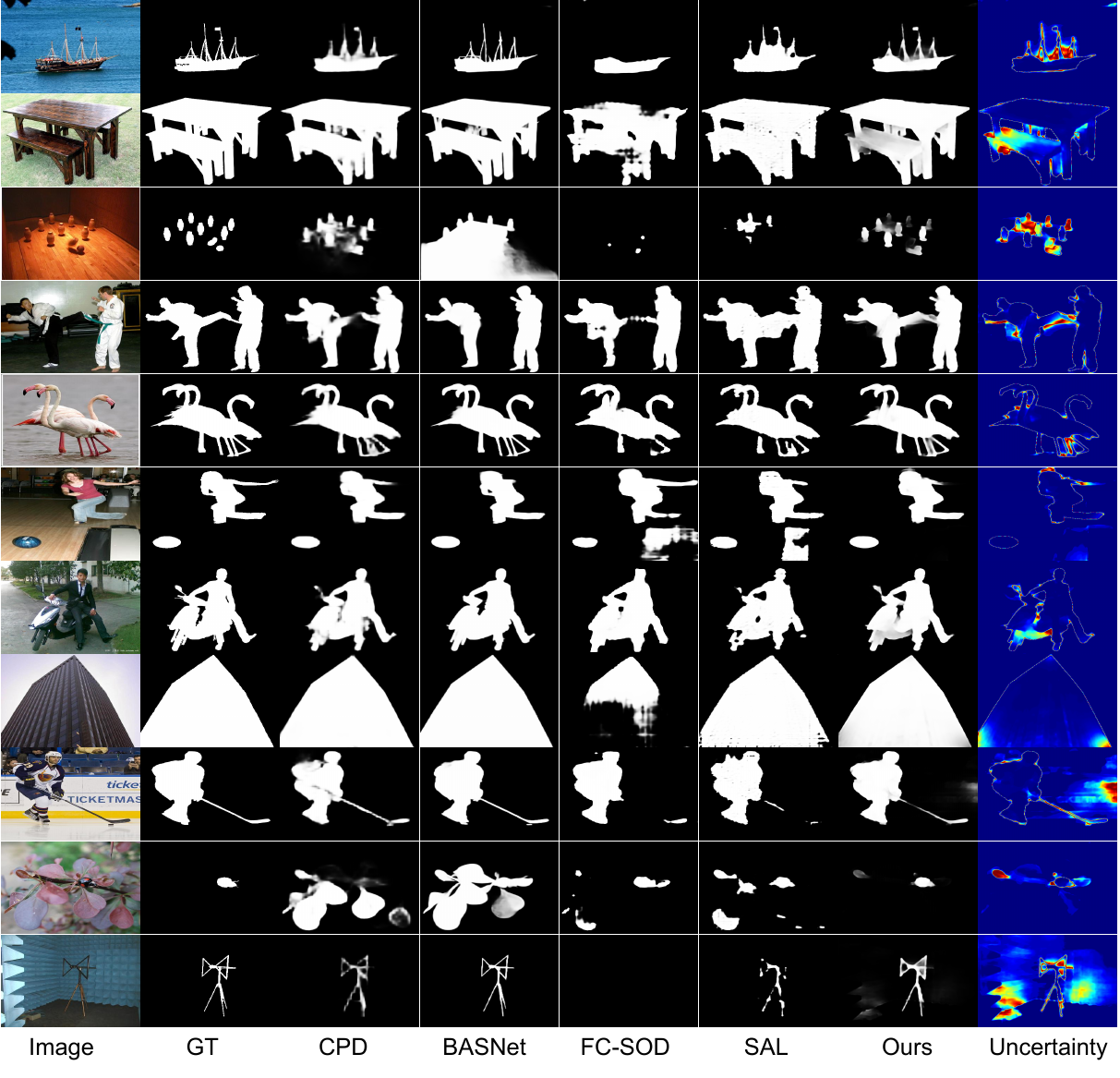}
    \caption{Predictions of semi-supervised (FC-SOD, SAL) and fully-supervised (CPD, BASNet) SOD methods, and our proposed method. Our results are associated with uncertainty maps highlighting predictions with low model confidence.}
    \label{fig:qualitative_results}
\end{figure*}

\begin{table*}[tbh!]
  \centering
  \scriptsize
  \renewcommand{\arraystretch}{1.2}
  \renewcommand{\tabcolsep}{4.0mm}
  \caption{
      Model performance ($F_{\xi}^{\mathrm{max}}$) w.r.t.~the usage of unlabeled set
      (\enquote{Base} vs \enquote{Full}) with only 659 ($\frac{1}{16}$) labeled data.
  }
  \begin{tabular}{lcccccc}
      \toprule
      Model & DUTS-TE & DUT-OMRON & PASCAL-S & SOD & ECSSD & HKU-IS\\
      \midrule
      Base & 0.840 $\pm$ 0.011 & 0.748 $\pm$ 0.006 & 0.846 $\pm$ 0.006 & 0.840 $\pm$ 0.007 & 0.919 $\pm$ 0.004 & 0.915 $\pm$ 0.003\\
      Full & \textbf{0.869 $\pm$ 0.004} & \textbf{0.774 $\pm$ 0.005} & \textbf{0.864 $\pm$ 0.004} & \textbf{0.856 $\pm$ 0.004} & \textbf{0.931 $\pm$ 0.001} & \textbf{0.928 $\pm$ 0.001}\\
      \bottomrule
  \end{tabular}
  \label{tab:different_labeled_set_splits}
\end{table*}

\subsection{Performance Comparison}
\noindent\textbf{Quantitative Comparison:}
In Tab.~\ref{tab:SS-SOD_benchmark}, we show performance of fully-supervised, weakly-supervised and semi-supervised SOD models for comparison.
It can be seen in Tab.~\ref{tab:SS-SOD_benchmark} that our proposed method performs favourably against previous state-of-the-art semi-supervised SOD methods on all six datasets. It is worth noting that we achieve these performances using only 659 fully labeled images, which is much less compared to the 1,000 images required by SaliencyGAN \cite{SaliencyGAN} and FC-SOD \cite{FC-SOD}, and the 1,500 images used in SAL \cite{SAL}. We significantly outperform FC-SOD and SAL in terms of the F-max measure on the DUTS, PASCAL-S, ECSSD and HKU-IS datasets. Further, we use a smaller ResNet-50 backbone compared to the ResNet-101 backbone used in FC-SOD. Our mean absolute errors $\mathcal{M}$ are comparable to SAL. This can be attributed to the fact that we use far fewer pixel-level labels, resulting in smoothed, though correct, predictions. 
With only 659 fully labeled images, the proposed method outperforms the SOD methods trained with weak annotations including image-level labels, scribbles and captions. Tab.~\ref{tab:SS-SOD_benchmark} also shows that our approach achieves comparable performance with fully-supervised SOD techniques.

\noindent\textbf{Qualitative Comparison:}
In Fig.~\ref{fig:qualitative_results}, we compare the SOD predictions of our method against existing semi-supervised and fully-supervised SOD methods. It can be observed that FC-SOD \cite{FC-SOD} often misses the target object partially or even entirely ($1^{\text{st}}$, $5^{\text{th}}$, $8^{\text{th}}$, $9^{\text{th}}$, $11^{\text{th}}$ rows). On the other hand, SAL \cite{SAL} tends to make bulky predictions. They are less likely to miss the salient object, compared to FC-SOD, however, the fine details around the structure boundary are lost, ($1^{\text{st}}$, $2^{\text{nd}}$, $4^{\text{th}}$ rows). Our model is more accurate in locating the target object, avoiding severe under- or over-segmentation. Moreover, the predictions of our model retain the structural details of complicated objects, e.g., the masts of the boat (row 1), the mirror of the motorcycle (row 7), the legs of the cranes (row 5), the boundaries of the bench (row 2), etc. Our method also presents the uncertainties associated with its predictions. They clearly indicate the complex area where predictions are prone to mistakes. Surprisingly, the uncertainty map of the last row even captures the shadow of the arm and the leg, which are difficult to distinguish. Samples in rows 2 and 4 also present the cause of reduced improvements in MAE with respect to the F-max measure. As our model is trained with only 659 labeled samples and soft pseudo labels on unlabeled samples, it tends to make over-smoothed predictions, although these pixels can be correctly recalled. 
As an auxiliary output from the proposed stochastic prediction network, Fig.~\ref{fig:qualitative_results} shows that the uncertainties associated with its predictions is reliable in explaining the less accurate predictions. \eg~the uncertainty map of the $4^{\text{th}}$ row even captures the silhouette of the arm and the leg, which are difficult to distinguish. The sample in the $2^{\text{nd}}$ and $4^{\text{th}}$ rows also presents the cause (smoothed predictions) of less significant improvement in MAE than in the F-max measure.

\subsection{Ablation Study}
\noindent\textbf{Semi-supervised learning:}
We evaluate the baseline performance on the 1/16 data split ratio. The model trained with only labeled data is denoted as the \enquote{Base} in Tab.~\ref{tab:different_labeled_set_splits}. \enquote{Full} represents our semi-supervised model (\enquote{Ours} in Tab.~\ref{tab:SS-SOD_benchmark}).
We conduct the experiments with five randomly labeled set splits and report and mean and standard deviation of the performance.
Tab.~\ref{tab:different_labeled_set_splits} shows that the \enquote{Base} model is sensitive to the initial labeled set selection, leading to larger performance standard deviation than \enquote{Full}.
Further, \enquote{Full} consistently outperforms \enquote{Base} on $F_{\xi}^{\mathrm{max}}$, indicating the effectiveness of our semi-supervised learning model.

\begin{table*}[tbh!]
    \centering
    \footnotesize
    \renewcommand{\arraystretch}{1.3}
    \renewcommand{\tabcolsep}{3.0mm}
    \caption{Model performance ($F_{\xi}^{\mathrm{max}}$) \wrt~the usage of confidence-aware learning (\enquote{No\_Conf} vs \enquote{Full}), the prior distribution (\enquote{NP} vs \enquote{Full}) and the amount of labeled samples under fixed data split of 1/16 (\enquote{Ours\_(164/329) vs \enquote{Full}}).
    }
    \resizebox{\textwidth}{!}{\begin{tabular}{lccccccccccccc}
    \toprule
    \multirow{2}{*}{Methods} & \multirow{2}{*}{Label}& \multicolumn{2}{c}{DUTS-TE} & \multicolumn{2}{c}{DUT-OMRON} & \multicolumn{2}{c}{PASCAL-S} & \multicolumn{2}{c}{SOD} & \multicolumn{2}{c}{ECSSD} & \multicolumn{2}{c}{HKU-IS}\\
    & & $F_{\xi}^{\mathrm{max}}\uparrow$ & $\mathcal{M}\downarrow$ 
    & $F_{\xi}^{\mathrm{max}}\uparrow$ & $\mathcal{M}\downarrow$ 
    & $F_{\xi}^{\mathrm{max}}\uparrow$ & $\mathcal{M}\downarrow$ 
    & $F_{\xi}^{\mathrm{max}}\uparrow$ & $\mathcal{M}\downarrow$ 
    & $F_{\xi}^{\mathrm{max}}\uparrow$ & $\mathcal{M}\downarrow$ 
    & $F_{\xi}^{\mathrm{max}}\uparrow$ & $\mathcal{M}\downarrow$ \\
    \midrule
    No\_Conf & 659 ($\frac{1}{16}$) & 0.853 & 0.051 & 0.754 & 0.076 & 0.853 & 0.077 & 0.846 & 0.108 & 0.923 & 0.047 & 0.919 & 0.035\\
    NP & 659 ($\frac{1}{16}$) & 0.859 & 0.050 & 0.764 & 0.076 & 0.856 & 0.072 & 0.849 & 0.109 & 0.926 & 0.046 & 0.924 & 0.035\\
    Ours\_164 & 164 ($\frac{1}{16}$) & 0.851 & 0.059 & 0.758 & 0.083 & 0.851 & 0.079 & 0.837 & 0.113 & 0.918 & 0.052 & 0.918 & 0.040\\
    Ours\_329 & 329 ($\frac{1}{16}$) & 0.856 & 0.056 & 0.765 & 0.081 & 0.856 & 0.077 & 0.842 & 0.108 & 0.927 & 0.047 & 0.924 & 0.036\\
    \hline
    Full & \textbf{659 ($\frac{1}{16}$)} & \textbf{0.869} & \textbf{0.045} & \textbf{0.774} & \textbf{0.067} & \textbf{0.864} & \textbf{0.067} & \textbf{0.856} & \textbf{0.104} & \textbf{0.931} & \textbf{0.043} & \textbf{0.928} & \textbf{0.034} \\
    \bottomrule
    \end{tabular}
    }
    \label{tab:ablation_study_on_confidence_estiomation_and_fixed_split_ratio}
\end{table*}

\noindent\textbf{Latent variable with Gaussian prior:} To test how the energy-based prior contributes to the proposed semi-supervised SOD model,
we train \enquote{Full} in Tab.~\ref{tab:different_labeled_set_splits} with
latent variables sampled
directly from the Gaussian distribution for both training and inference, and denote its performance as \enquote{NP}. Tab.~\ref{tab:different_labeled_set_splits} shows its deteriorated performance compared with the \enquote{Full} model,
with the largest gap observed DUT-OMRON dataset. It shows that the data distribution approximated by the latent variable is more meaning than the Gaussian distribution assumption.

\noindent\textbf{Uncertainty estimation:} We compare the models (1) without uncertainty estimation (\enquote{No\_Conf}) with uncertainty estimation, and (2) with uncertainty estimation (\enquote{Full}), in the learning of unlabeled data in Tab.~\ref{tab:ablation_study_on_confidence_estiomation_and_fixed_split_ratio}. It demonstrates that the proposed uncertainty estimation is accurate in locating the potentially incorrect areas in the pseudo labels, as illustrated in Fig.~\ref{fig:qualitative_results}, preventing our model from learning this noise. This is reflected in the model performances where \enquote{Full} consistently outperforms \enquote{No\_Conf} on all 6 testing datasets.



\begin{figure*}[htb!]
    \centering
    \begin{tabular}{{c@{ } c@{ } c@{ }}}
        {\includegraphics[width=0.32\linewidth]{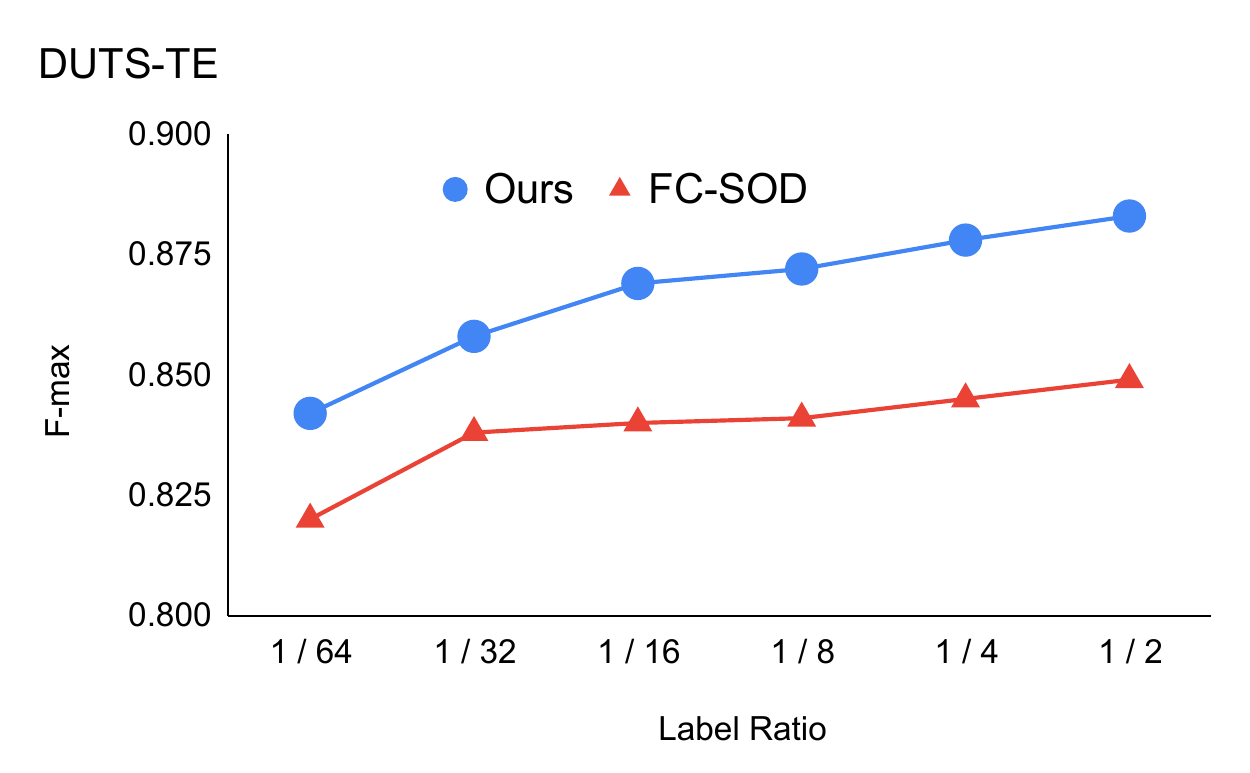}} &
        {\includegraphics[width=0.32\linewidth]{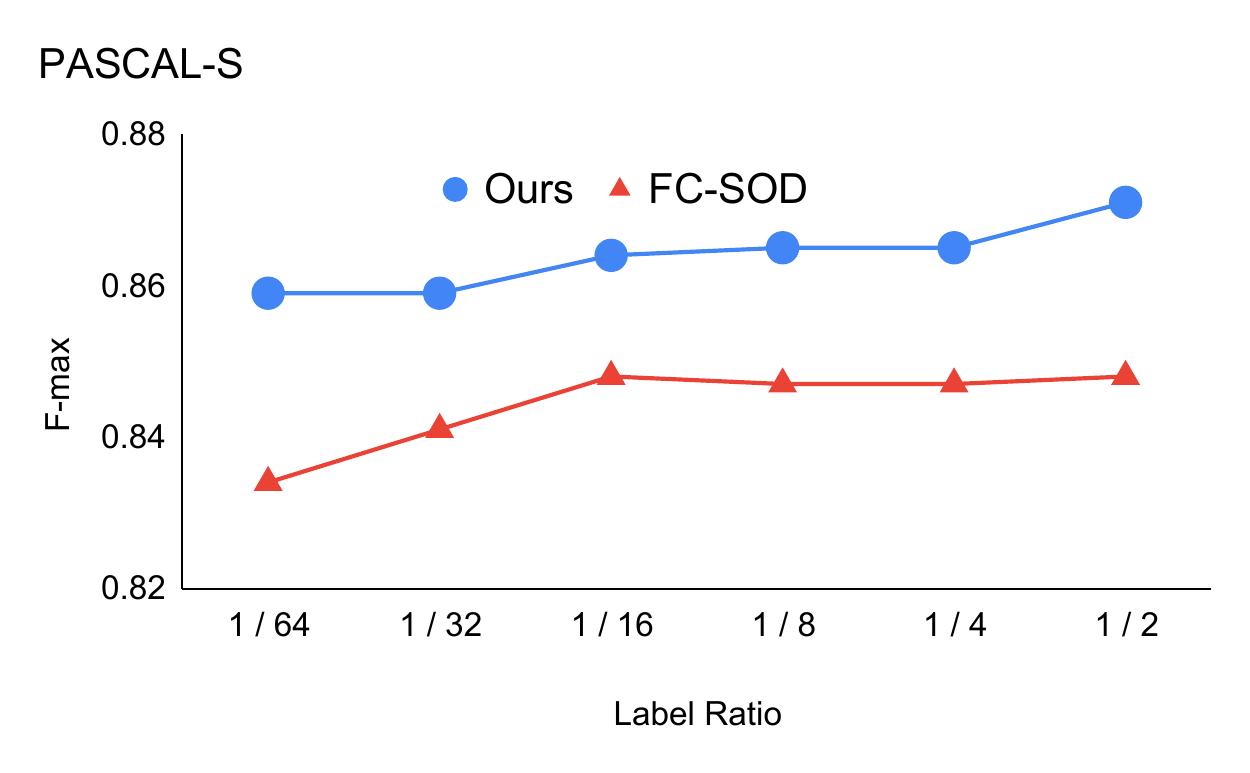}} &
        {\includegraphics[width=0.32\linewidth]{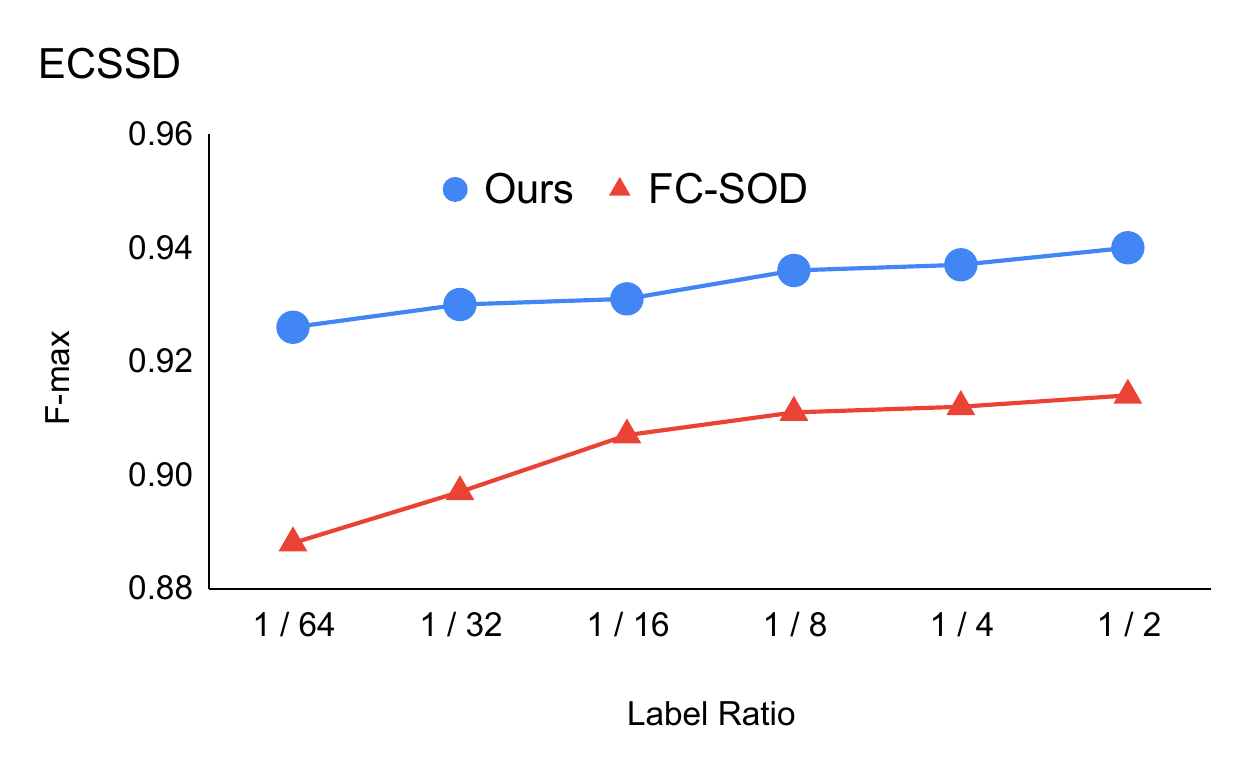}}
    \end{tabular}
    \begin{tabular}{{c@{ } c@{ } c@{ }}}
        {\includegraphics[width=0.32\linewidth]{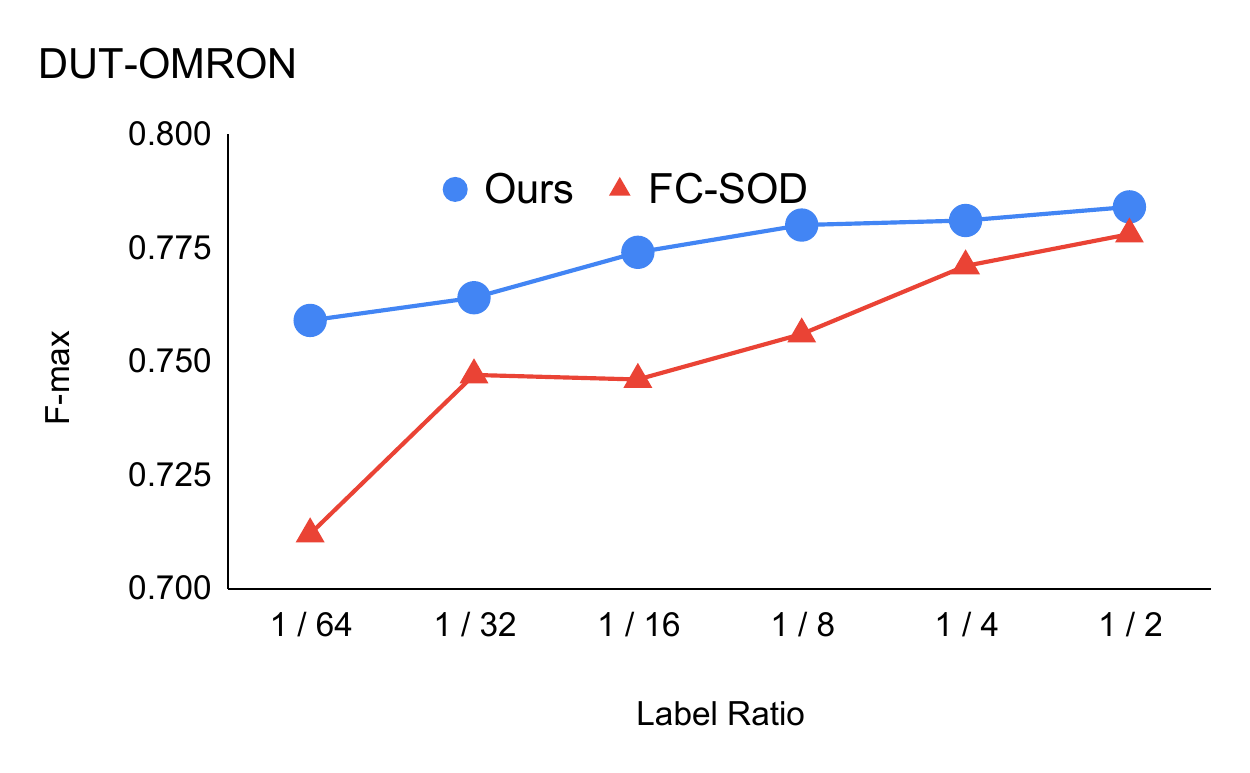}} &
        {\includegraphics[width=0.32\linewidth]{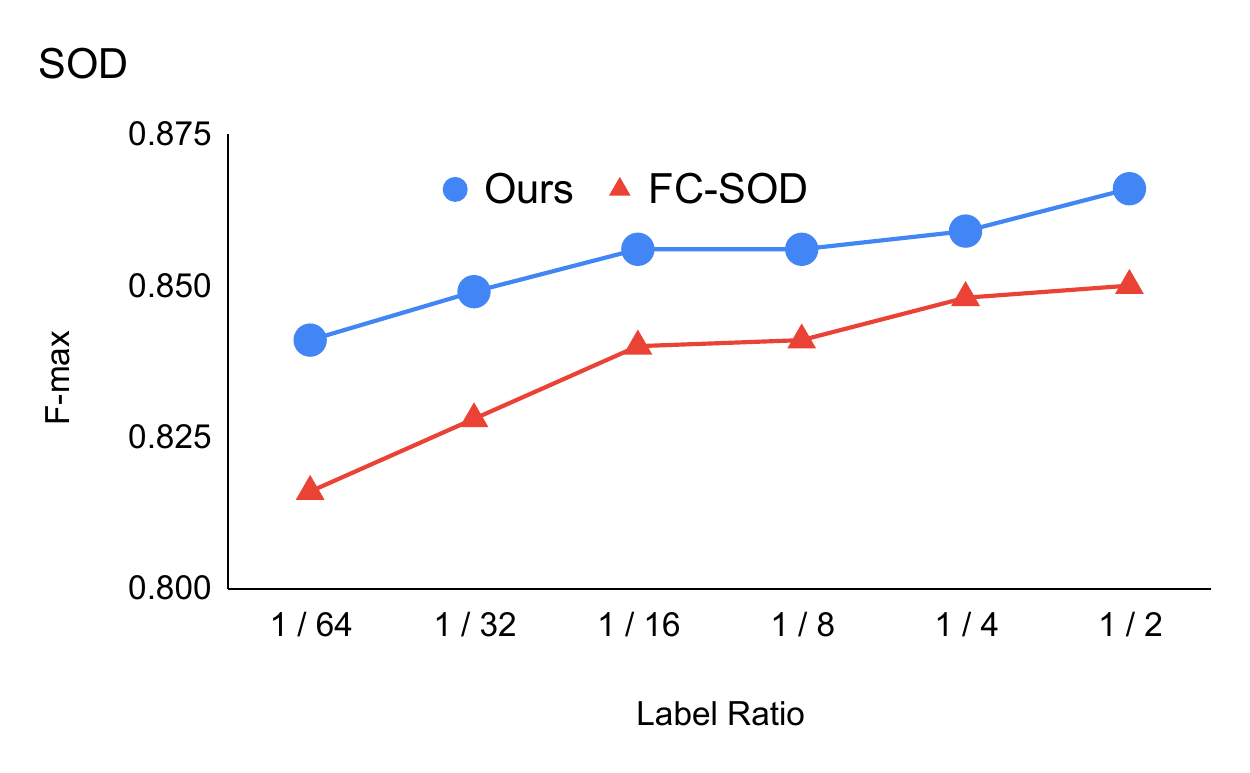}} &
        {\includegraphics[width=0.32\linewidth]{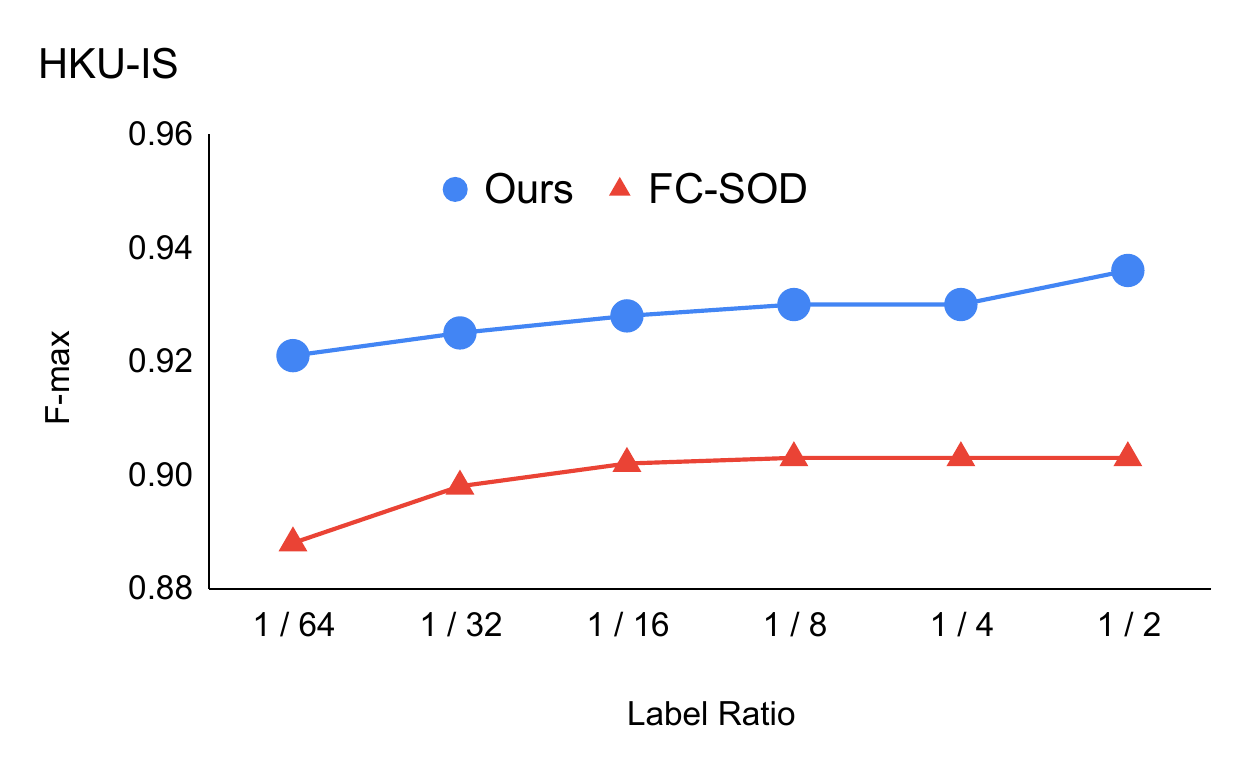}}
    \end{tabular}
    \caption{
        Comparisons of baseline, our model and FC-SOD trained on different dataset splits including 1/64, 1/32, 1/16, 1/8, 1/4 and 1/2 on the six benchmark testing datasets, DUTS-TE \cite{DUTS-TE}, DUT-OMRON \cite{DUT-OMRON}, PASCAL-S \cite{PASCAL-S}, SOD \cite{SOD}, ECSSD \cite{ECSSD}, HKU-IS \cite{li2015visual}. (Best viewed in colour)
    }
    \label{fig:ablation_study_different_dataset_splits}
\end{figure*}

\noindent\textbf{Model performance with respect to labeled set ratio:}
Fig.~\ref{fig:ablation_study_different_dataset_splits} compares our \enquote{Full} model with FC-SOD \cite{FC-SOD} trained with six different labeled set ratios including 1/64, 1/32, 1/16, 1/8, 1/4 and 1/2 using the F-max measure. It illustrates that our method consistently outperforms FC-SOD on all labeled data ratios. More significant improvements can be observed at low data regimes. Our model trained with 1/64 of the labeled data even outperforms FC-SOD trained with 1/2 on PASCAL-S, ECSSD and HKU-IS. We also observe a significant performance lead of our model with 1/32 labeled data over FC-SOD with 1/2 labeled data on DUTS-TE dataset. FC-SOD has strong performances on DUTS-OMRON and SOD datasets. However, conditioned on the same labeled set ratio, our \enquote{Full} model still achieves better results.


\begin{table*}[htb!]
    \centering
    \scriptsize
    \renewcommand{\arraystretch}{1.3}
    \renewcommand{\tabcolsep}{4.5mm}
    \caption{Performance comparison between our model and the alternative deterministic and generative model based semi-supervised solutions on the benchmark testing datasets. * indicates our re-implementation.}
    \resizebox{\textwidth}{!}{\begin{tabular}{lcccccccccccc}
        \toprule
        \multirow{2}{*}{Model}& \multicolumn{2}{c}{DUTS-TE} & \multicolumn{2}{c}{DUT-OMRON} & \multicolumn{2}{c}{PASCAL-S} & \multicolumn{2}{c}{SOD} & \multicolumn{2}{c}{ECSSD} & \multicolumn{2}{c}{HKU-IS}\\
        & $F_{\xi}^{\mathrm{max}}\uparrow$ & $\mathcal{M}\downarrow$ & $F_{\xi}^{\mathrm{max}}\uparrow$ & $\mathcal{M}\downarrow$ & $F_{\xi}^{\mathrm{max}}\uparrow$ & $\mathcal{M}\downarrow$ & $F_{\xi}^{\mathrm{max}}\uparrow$ & $\mathcal{M}\downarrow$ & $F_{\xi}^{\mathrm{max}}\uparrow$ & $\mathcal{M}\downarrow$ & $F_{\xi}^{\mathrm{max}}\uparrow$ & $\mathcal{M}\downarrow$ \\
        \midrule
        Base* & 0.840 & 0.058 & 0.748 & 0.090 & 0.850 & 0.075 & 0.829 & 0.125 & 0.917 & 0.052 & 0.911 & 0.043\\
        $\Pi$ Model* & 0.831 & 0.065 & 0.747 & 0.088 & 0.844 & 0.090 & 0.833 & 0.120 & 0.915 & 0.057 & 0.915 & 0.044 \\
        MT* & 0.853 & 0.048 & 0.760 & 0.074 & 0.857 & 0.072 & 0.841 & 0.113 & 0.925 & 0.049 & 0.922 & 0.036\\
        \hline
        Ours & \textbf{0.869} & \textbf{0.045} & \textbf{0.774} & \textbf{0.067} & \textbf{0.864} & \textbf{0.067} & \textbf{0.856} & \textbf{0.104} & \textbf{0.931} & \textbf{0.043} & \textbf{0.928} & \textbf{0.034} \\
        \bottomrule
  \end{tabular}}
  \label{tab:comparison_with_alternative_semi-supervised_SOD_solutions}
\end{table*}

\noindent\textbf{Entropy Loss:}
The Entropy Loss prevents the model, which optimises is optimised on unlabeled samples with soft pseudo labels, from outputting overly smoothed predictions. We ablate the Entropy Loss by setting $\lambda_{\text{ue}} = 0$. As shown in Tab.~\ref{tab:ablation_study_on_entropy_loss}, when trained on the unlabeled samples without the Entropy Loss, the model produces inferior performance in terms of Mean Absolute Error ($\mathcal{M}$ metric).

\begin{table*}[t!]
  \centering
  \scriptsize
  \renewcommand{\arraystretch}{1.3}
  \renewcommand{\tabcolsep}{1.3mm}
  \caption{The effect of Entropy Loss on CEBM}
  \begin{tabular}{lc|cc|cccccccccccc}
  \toprule
  \multirow{2}{*}{Methods} & 
  \multirow{2}{*}{Label} & \multicolumn{2}{c|}{Loss Component} & \multicolumn{2}{c}{DUTS-TE} & \multicolumn{2}{c}{DUT-OMRON} & \multicolumn{2}{c}{PASCAL-S} & \multicolumn{2}{c}{SOD} & \multicolumn{2}{c}{ECSSD} & \multicolumn{2}{c}{HKU-IS}\\
  & & $\mathcal{L}_{\text{str}}$ & 
  $\mathcal{L}_{\text{ent}}$
  & $F_{\xi}^{\mathrm{max}}\uparrow$ & $\mathcal{M}\downarrow$ & $F_{\xi}^{\mathrm{max}}\uparrow$ & $\mathcal{M}\downarrow$ & $F_{\xi}^{\mathrm{max}}\uparrow$ & $\mathcal{M}\downarrow$ & $F_{\xi}^{\mathrm{max}}\uparrow$ & $\mathcal{M}\downarrow$ & $F_{\xi}^{\mathrm{max}}\uparrow$ & $\mathcal{M}\downarrow$ & $F_{\xi}^{\mathrm{max}}\uparrow$ & $\mathcal{M}\downarrow$ \\
  \midrule
  CEBM & \textbf{659 ($\frac{1}{16}$)} & \checkmark & - & 0.869 & 0.050 & 0.773 & 0.073 & 0.863 & 0.075 & 0.857 & 0.112 & 0.930 & 0.046 & 0.928 & 0.038 \\
  CEBM & \textbf{659 ($\frac{1}{16}$)} & \checkmark & \checkmark & 0.869 & 0.045 & 0.774 & 0.067 & 0.864 & 0.067 & 0.856 & 0.104 & 0.931 & 0.043 & 0.928 & 0.034 \\
  \bottomrule
  \end{tabular}
  \label{tab:ablation_study_on_entropy_loss}
\end{table*}

\subsection{Model Discussion}
\begin{figure*}[htb!]
    \centering
    \includegraphics[width=\linewidth]{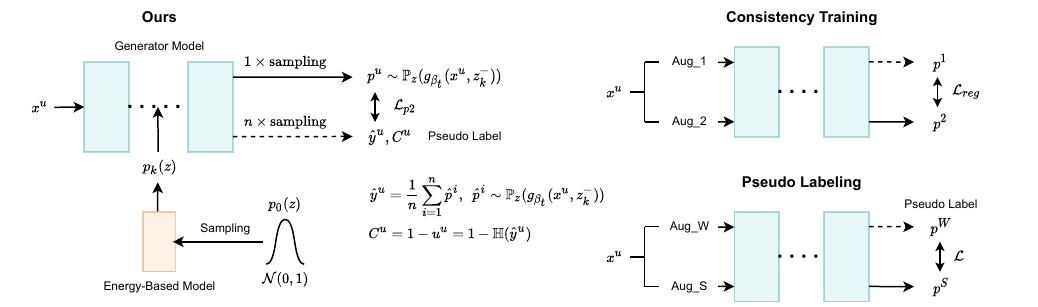}
    \caption{Illustrations of semi-supervised learning with the consistency regularisation based method and pseudo-label based method. The primary difference between these two methods lies in the utilisation of the unlabeled data during the training.
    Our proposed method belongs to the second category. 
    We contribute a conditional energy based model and a confidence-aware learning enabled by the stochastic attribute the generative model. Loss is only back-propagated through solid arrow line.}
    \label{fig:semi_supervised_learning_methods}
\end{figure*}

\noindent\textbf{Alternative deterministic semi-supervised solutions:}
Semi-Supervised Learning (SSL) addresses the annotation shortage by incorporating a large amount of unlabeled data in training a classifier. It has been extensively studied for classification tasks with many works focusing on consistency based \cite{ladder_network,pi_model,temporal_ensemble,mean_teacher,virtual_adversarial_training,unsupervised_data_augmentation} and pseudo-label based \cite{co-training,pseudo_label,noisy_student,S4L,EnAET} techniques (see Fig.~\ref{fig:semi_supervised_learning_methods}). Both of these methods initiate the training on the labeled data and differ 
on the unlabeled part. Consistency based methods constrain the model to produce similar predictions for unlabeled training samples under different data augmentation schemes. Popular data augmentation techniques include rotation, flipping, scaling, translation, cropping, colour jittering, etc. The pseudo-label based methods usually perform semi-supervised learning following a three-step learning approach, including 1) training with the labeled set; 2) generating pseudo labels for the unlabeled set; 3) training further using the unlabeled set with the corresponding pseudo labels. We re-implement them with our generator model for the semi-supervised SOD task.

\begin{figure*}[htb!]
    \centering
    \includegraphics[width=\textwidth]{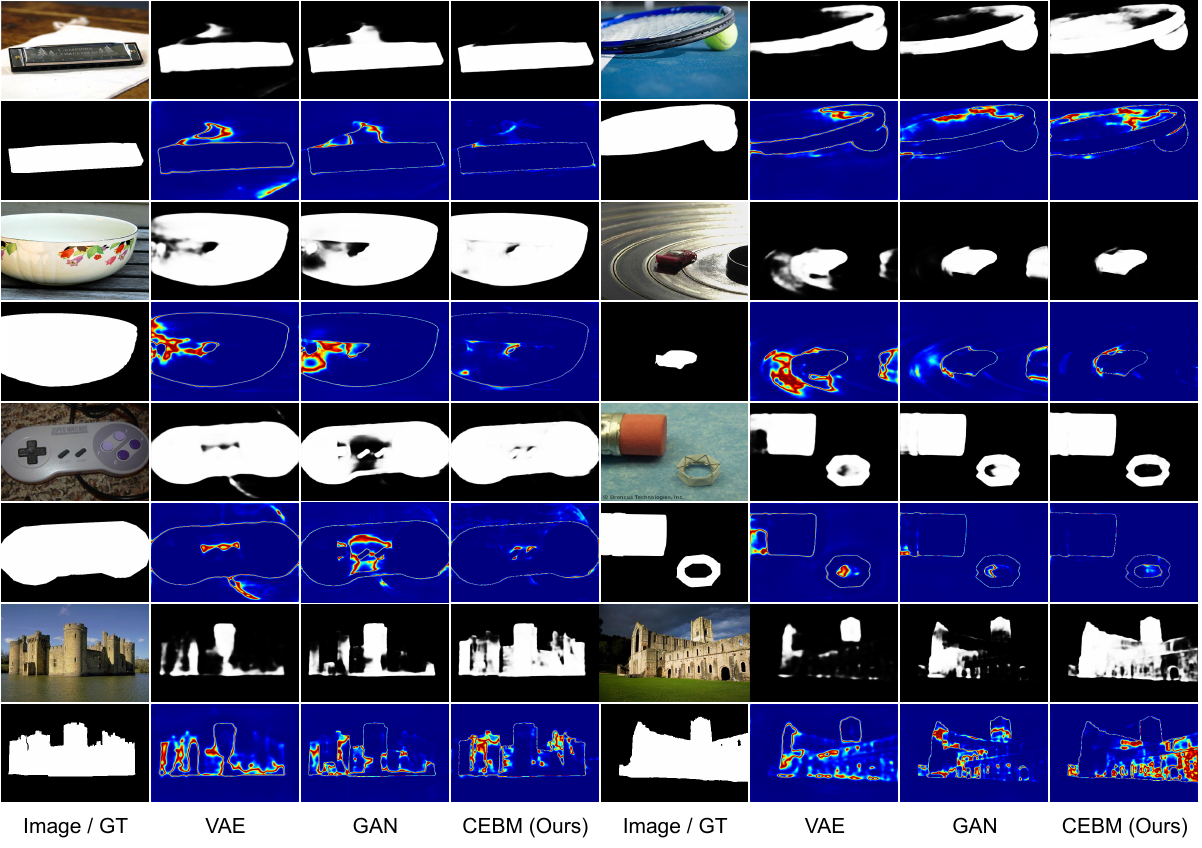}
    \caption{Pseudo-labels and their associated uncertainty maps of the unlabeled sample estimated by three generative models including VAE, GAN and CEBM (Ours).}
    \label{fig:supplementary_generative_model_comparison}
\end{figure*}

\textit{$\Pi$ model \cite{pi_model} for semi-supervised SOD:} The \enquote{$\Pi$ Model*} enforces consistent predictions under different data augmentations, e.g., scaling, flipping and rotation. To adapt it for our semi-supervised task,
we adopt a typical U-Net structure with a ResNet50 encoder and the decoder of MiDaS \cite{MiDaS}. To enforce consistent predictions,
each training image is forward propagated twice through the network under different combinations of scaling, rotation and flipping. The training is single-phase, involving both labeled and unlabeled images. The total loss is defined as $\mathcal{L} = \mathcal{L}_{sup} + \mathcal{L}_{con}$, where $\mathcal{L}_{sup}$ is a binary cross-entropy loss on the labeled data and $\mathcal{L}_{con}$ is an L2 loss applied to both labeled and unlabeled data.

\textit{Mean Teacher (MT) for semi-supervised SOD:} MT \cite{mean_teacher} has a teacher model and a student model, each adopting a U-Net structure with a ResNet encoder and the decoder of MiDaS \cite{MiDaS}. The teacher model adopts the moving average of the weight of the student model as defined in:
\begin{equation}
    \theta_{t}^{T} = \alpha \theta_{t-1}^{T} + (1 - \alpha) \theta_{t}^{S},
\end{equation}
where $T$ and $S$ indicate the teacher and the student model, $t$ denotes the current training epoch and $\alpha$ is a hyperparameter that we empirically set to 0.95. Consistency is enforced between predictions of the teacher and the student model for each input image.

The performance of these models trained with the 1/16 data split ratio are presented in Tab.~\ref{tab:comparison_with_alternative_semi-supervised_SOD_solutions}. Here, \enquote{Base*} model is a saliency prediction network same as \enquote{Base} in Tab.~\ref{tab:different_labeled_set_splits}.
\enquote{$\Pi$ Model*} does not show much improvement over the \enquote{Base*} model. \enquote{MT*} is a more sophisticated semi-supervised learning model for classification. The results show its efficacy in semi-supervised SOD, achieving consistent improvements over the \enquote{Base*} model. However, our model still outperforms \enquote{MT*} on all testing datasets.

\noindent\textbf{Comparison with alternative generative model based solutions:} 
We compare our solution with existing generative solutions from two perspectives, including pseudo-label quality and uncertainty map reliability.

\textit{Pseudo-Label Quality:}
Pseudo-labels of the unlabeled samples obtained from the three generative solutions, including VAE, GAN and CEBM (ours), are presented in Fig.~\ref{fig:supplementary_generative_model_comparison}. 
It can be observed that our method produces the highest-quality masks on the unlabeled samples. Especially in the hard samples (\enquote{tennis racquet}, \enquote{castle}), GAN and VAE miss a significant part of the target objects, but our model yields more holistic pseudo-labels.

\textit{Uncertainty Map:}
Fig.~\ref{fig:supplementary_generative_model_comparison} depicts the uncertainty maps of the three generative model based solutions. It can be seen that uncertainty maps of our CEBM are able to highlight the regions with low model confidence, where the pixel-wise pseudo-labels have a high probability of being incorrect. The confidence-aware learning allows our model to ignore those potentially false pseudo-labels, preventing the persistent error propagation issue. On the other hand, the uncertainty maps of GAN and VAE are less precise. While their pseudo-labels miss a significant part of \enquote{tennis racquet} and \enquote{castle}, their uncertainty maps fail to associate those incorrect labels with low confidence.

\begin{table*}[htb!]
    \centering
    \footnotesize
    \renewcommand{\arraystretch}{1.3}
    \renewcommand{\tabcolsep}{2.0mm}
    \caption{Large-scale experiment with our proposed CEBM using DUTS-TE as labeled set and PASCAL-VOC 2012 \cite{everingham2010pascal} as unlabeled set.}
    \resizebox{\textwidth}{!}{\begin{tabular}{lcccccccccccccc}
        \toprule
        \multirow{2}{*}{Model} & \multicolumn{2}{c}{Data} & \multicolumn{2}{c}{DUTS-TE} & \multicolumn{2}{c}{DUT-OMRON} & \multicolumn{2}{c}{PASCAL-S} & \multicolumn{2}{c}{SOD} & \multicolumn{2}{c}{ECSSD} & \multicolumn{2}{c}{HKU-IS}\\
        & Labeled & Unlabeled &
        $F_{\xi}^{\mathrm{max}}\uparrow$ & $\mathcal{M}\downarrow$ & $F_{\xi}^{\mathrm{max}}\uparrow$ & $\mathcal{M}\downarrow$ & $F_{\xi}^{\mathrm{max}}\uparrow$ & $\mathcal{M}\downarrow$ & $F_{\xi}^{\mathrm{max}}\uparrow$ & $\mathcal{M}\downarrow$ & $F_{\xi}^{\mathrm{max}}\uparrow$ & $\mathcal{M}\downarrow$ & $F_{\xi}^{\mathrm{max}}\uparrow$ & $\mathcal{M}\downarrow$ \\
        \midrule
        $\text{CEBM}_{\text{D}}$ & 10,553 & - & 0.885 & 0.040 & 0.808 & 0.055 & 0.873 & 0.066 & 0.871 & 0.093 & 0.943 & 0.040 & 0.936 & 0.033\\
        $\text{CEBM}_{\text{D \& P}}$ & 10,553 & 10,582 & \textbf{0.897} & \textbf{0.035} & \textbf{0.821} & \textbf{0.053} & \textbf{0.881} & \textbf{0.060} & 0.875 & \textbf{0.092} & \textbf{0.949} & \textbf{0.033} & \textbf{0.944} & \textbf{0.027} \\
        \bottomrule
  \end{tabular}}
  \label{tab:large_scale_experiments}
\end{table*}

\begin{figure*}
    \centering
    \includegraphics[width=\textwidth]{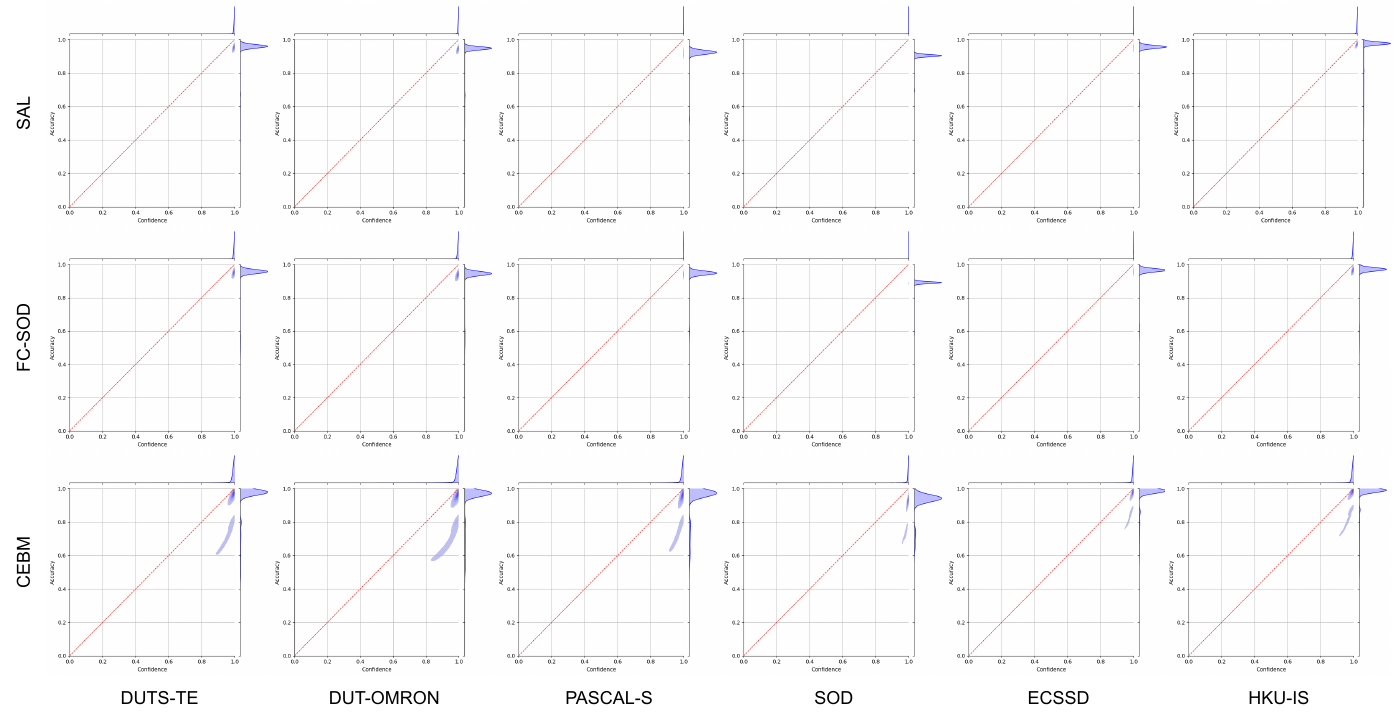}
    \caption{Histogram plot of the prediction distribution (after sigmoid) of the existing semi-supervised SOD models (FC-SOD \cite{FC-SOD} and SAL \cite{SAL}) and our proposed method}
    \label{fig:histogram_prediction_distribution}
\end{figure*}


\noindent\textbf{Large-scale experiment:}
We conduct a large-scale experiment by using the whole DUTS training set (10,533 images) as the labeled set and the augmented PASCAL VOC 2012 training set (10,582 images) \cite{everingham2010pascal} as the unlabeled set. The resultant labeled split ratio is 1/2. We denote the model trained with only the labeled part as (\enquote{$\text{CEBM}_{\text{D}}$}) and the full model as (\enquote{$\text{CEBM}_{\text{D \& P}}$}). Tab.~\ref{tab:large_scale_experiments} shows that our model trained with the whole DUTS training set in a fully-supervised fashion ($\text{CEBM}_{\text{D}}$) achieves similar performance to existing state-of-the-art SOD models. Further, the incorporation of additional unlabeled data into our confidence-aware semi-supervised learning further boosts the performance on all benchmark testing datasets. The final model ($\text{CEBM}_{\text{D\&P}}$) can outperform the existing fully-supervised models with exception only on the SOD \cite{SOD} dataset, where EGNet \cite{EGNet} has better performance in terms of maximum F-measure. However, our mean absolute error score is better in this case. The experiment demonstrates that our proposed method is also effective in improving the model over the existing SOD models with the aid of additional unlabeled data.

\noindent\textbf{Comparison with existing semi-supervised salient object detection (SS-SOD) methods in terms of model calibration:}
We adopt the expected calibration error (ECE) \cite{guo2017calibration,liu2023model}, where we set the number of bins to 10, to measure the model calibration degrees of the semi-supervised SOD models. Tab.~\ref{tab:ece} shows the ECE results of SAL, FC-SOD and our ECBM on the xis testing datasets. CEBM has significantly improved over both SAL and FC-SOD on all testing datasets. Fig.~\ref{fig:histogram_prediction_distribution} presents the joint distribution of prediction confidence (horizontal axis) and prediction accuracy (vertical axis) of SAL \cite{SAL}, FC-SOD and our proposed CEBM. Both FC-SOD and SAL produce highly confident predictions and their joint distributions have almost shrunk to a point mass on all six testing datasets. On the other hand, the joint distribution of our CEBM is closer to the oracle line, which represents a perfectly calibrated model.

\begin{table}[htb!]
    \centering
    \scriptsize
    \renewcommand{\arraystretch}{1.3}
    \renewcommand{\tabcolsep}{3.0mm}
    \caption{Expected calibration error of SAL \cite{SAL}, FC-SOD \cite{FC-SOD} and our CEBM evaluated on DUTS-TE, DUT-OMRON, PASCAL-S, SOD, ECSSD, HKU-IS.}
    \begin{tabular}{lccc}
        \toprule
        \multirow{2}{*}{Dataset} & \multicolumn{3}{c}{ECE (\%)} \\
        \cmidrule{2-4}
        & SAL & FC-SOD & CEBM\\
        \midrule
        DUTS-TE & 5.01 & 4.71 & 4.18\\
        DUT-OMRON & 6.25 & 5.76 & 5.31\\
        PASCAL-S & 6.02 & 8.28 & 5.20\\
        SOD & 11.58 & 9.99 & 8.22\\
        ECSSD & 4.20 & 4.89 & 2.60\\
        HKU-IS & 3.50 & 2.71 & 2.04\\
        \bottomrule
    \end{tabular}
    \label{tab:ece}
\end{table}

Fig.~\ref{fig:histogram_prediction_distribution} compares the prediction distributions between the existing semi-supervised SOD methods (FC-SOD \cite{FC-SOD} and SAL \cite{SAL}) and our proposed method. Our method is better calibrated than FC-SOD, making correct, yet non-over-confident predictions (predictions in $(0.0, 0.2)$ and $(0.8, 1.0)$). On the contrary, FC-SOD overfits to the datasets, making more absolute predictions (0.0 and 1.0). Further, FC-SOD has more potentially incorrect predictions in the range $(0.4, 0.6)$. These two phenomena explains why our method achieves comparable mean absolute error performance with FC-SOD while significantly outperforming it on the F-max measure.

\noindent\textbf{Model size and inference speed:}
The proposed CEBM has 86.70 GFLOPs and an inference speed of 40.38 images per second. The inference speed is averaged with 5 runs over the 6 testing datasets. These results are close to those of FC-SOD \cite{FC-SOD} which has 78.11 GFLOPS and an inference speed of 44.24 images per second.

\subsection{Failure Samples}
\begin{figure*}
    \centering
    \includegraphics[width=\textwidth]{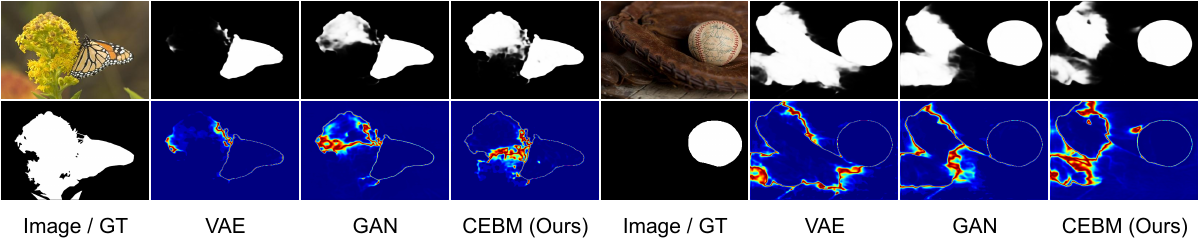}
    \caption{Failure examples of three generative models including VAE, GAN and CEBM (Ours).}
    \label{fig:failure_examples}
\end{figure*}

Fig.~\ref{fig:failure_examples} illustrates that generative models tend to produce inaccurate uncertainty maps when they have persistent yet incorrect predictions. However, our model suffers a less critical failure compared to GAN and VAE. The main reason lies in the missing divergence constraint of predictions with respect to the sampling of the latent space, which sometimes result in structural variance, but not enough semantic variance across the multiple predictions. We will work in this direction for more reliable confidence map generation.


\section{Conclusion}


We introduce a pseudo label based approach for semi-supervised salient object detection with a conditional energy-based model. We use the stochastic latent variable of CEBM to model the stochastic nature of human saliency labels, particularly from multiple annotators, and further compute the epistemic uncertainty on the pseudo labels with fine-grained details. This enables a confidence-aware semi-supervised learning framework that efficiently addresses the potential error propagation issue in the utilisation of unlabeled data. The proposed framework achieves favourable comparisons against the state-of-the-art SS-SOD models, and is even competitive to some fully-supervised SOD models.

\bibliographystyle{IEEEtran}
\bibliography{Reference}
\end{document}